\documentclass[10pt,twocolumn,letterpaper]{article}

\usepackage[pagenumbers]{cvpr}

\usepackage{epsfig}
\usepackage{graphicx}
\usepackage{amsmath}
\usepackage{amssymb}
\usepackage{pifont}

\usepackage{latexsym}
\usepackage{gensymb}
\usepackage{amsfonts}       
\usepackage{adjustbox}
\usepackage{multirow}
\usepackage{booktabs}
\usepackage{nicefrac}       
\usepackage{microtype}      
\usepackage[T1]{fontenc}
\usepackage[utf8]{inputenc} 
\usepackage{float}
\usepackage{makecell}
\usepackage{appendix}
\usepackage[hyphens]{url}
\usepackage{color,colortbl,xcolor}
\usepackage{caption}
\usepackage{algorithm}
\usepackage{listings}

\usepackage[accsupp]{axessibility}  

\definecolor{cvprblue}{rgb}{0.21,0.49,0.74}

\usepackage[pagebackref,breaklinks,bookmarks=false]{hyperref}
\hypersetup{
    colorlinks=true,
    linkcolor=red,
    citecolor=cvprblue,
    urlcolor=Green
}

\definecolor{Gray}{gray}{0.9}
\newcommand{\model}{SimLTD\xspace}
\newcommand{\lvis}{LVIS\xspace}
\newcommand{\dl}{$\mathcal{D}_\text{ltd}$\xspace}
\newcommand{\dtail}{$\mathcal{D}_\text{tail}$\xspace}
\newcommand{\dhead}{$\mathcal{D}_\text{head}$\xspace}
\newcommand{\dk}{$\mathcal{D}_k$\xspace}
\newcommand{\chead}{$C_\text{head}$\xspace}
\newcommand{\ctail}{$C_\text{tail}$\xspace}
\newcommand{\apbox}{AP$_\text{box}$\xspace}
\newcommand{\apr}{AP$_\text{r}$\xspace}
\newcommand{\apc}{AP$_\text{c}$\xspace}
\newcommand{\apf}{AP$_\text{f}$\xspace}
\newcommand{\xmark}{\ding{55}}

\usepackage[capitalize]{cleveref}
\crefname{section}{Sec.}{Secs.}
\Crefname{section}{Section}{Sections}
\Crefname{table}{Table}{Tables}
\crefname{table}{Tab.}{Tabs.}

\title{SimLTD: Simple Supervised and Semi-Supervised Long-Tailed Object Detection}

\author{
    Phi Vu Tran\\
    LexisNexis Risk Solutions
}

\begin{document}
\maketitle
\begin{abstract}
While modern visual recognition systems have made significant advancements, many continue to struggle with the open problem of learning from few exemplars. This paper focuses on the task of object detection in the setting where object classes follow a natural long-tailed distribution. Existing methods for long-tailed detection resort to external ImageNet labels to augment the low-shot training instances. However, such dependency on a large labeled database has limited utility in practical scenarios. We propose a versatile and scalable approach to leverage optional unlabeled images, which are easy to collect without the burden of human annotations. Our SimLTD framework is straightforward and intuitive, and consists of three simple steps: (1) pre-training on abundant head classes; (2) transfer learning on scarce tail classes; and (3) fine-tuning on a sampled set of both head and tail classes. Our approach can be viewed as an improved head-to-tail model transfer paradigm without the added complexities of meta-learning or knowledge distillation, as was required in past research. By harnessing supplementary unlabeled images, without extra image labels, SimLTD establishes new record results on the challenging LVIS v1 benchmark across both supervised and semi-supervised settings.
\end{abstract}
\section{Introduction}\label{sec::introduction}
The task of detecting, localizing, and classifying object instances from image and video is a long-standing problem in computer vision. Recent years have seen unprecedented progress on modern object detection systems, mostly driven by powerful neural architectures. Much of this success is measured on the relatively balanced, small-vocabulary benchmarks like PASCAL VOC~\cite{voc} and MS-COCO~\cite{coco}. When evaluated on a more complex and imbalanced dataset with a much larger vocabulary, however, the same models exhibit a considerable drop in detection accuracy~\cite{lvis}.

\begin{figure}[t]
\centering
\includegraphics[width=0.98\linewidth]{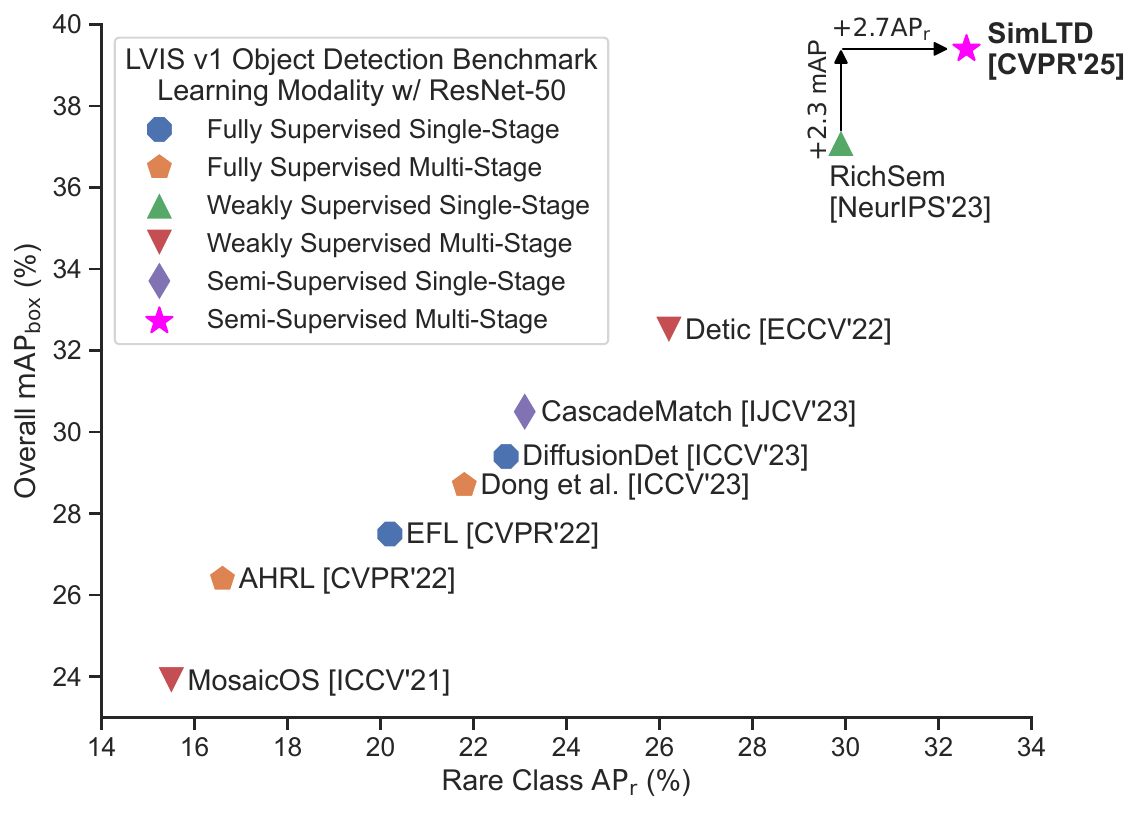}
\caption{A survey comparing our \model to the state of the art for long-tailed detection. We combine unlabeled data with an intuitive multi-stage training strategy to deliver the best overall performance while also optimizing for accuracy on rare classes. Our simple approach achieves superior results on the challenging \lvis v1 benchmark without requiring auxiliary image-level supervision.}
\label{fig::results-teaser}
\end{figure}
\begin{figure*}[t]
\centering
\includegraphics[width=0.9\textwidth]{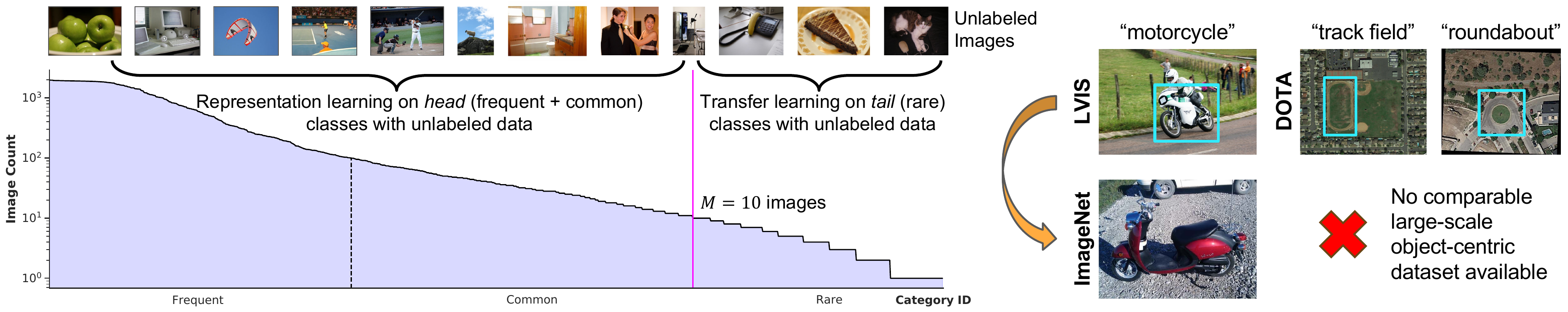}
\caption{The motivation to our approach. \textbf{Left:} We propose an improved head-to-tail model transfer framework for long-tailed detection by incorporating \emph{unlabeled images} in both representation and transfer learning stages. \textbf{Right:} While it is possible to find more samples of \lvis instances from ImageNet, such an auxiliary database may not exist in another scenario like aerial imagery. Our framework does not depend on using extra image-level labels to advance long-tailed detection, which expands the applicability of our approach beyond \lvis.}
\label{fig::motivation}
\end{figure*}

This paper explores ways to enhance the capability of commodity detection systems, with a particular evaluation emphasis on the challenging large-vocabulary \lvis benchmark~\cite{lvis}. \lvis represents a realistic application, a scenario in which object classes follow a natural long-tailed distribution. It is in the tail that most data-hungry models struggle with performance, a distribution characterized by many rare classes having as few as a single training exemplar.

\Cref{fig::results-teaser} plots several recent state-of-the-art methods to address the extreme disparity of class distributions in long-tailed detection (LTD). One promising direction is to divide the problem into multiple manageable parts to help alleviate the severity of the class imbalance. LST~\cite{lst} proposes to segment the overall dataset into seven phases, each phase containing progressively smaller but balanced data samples, and train a model in an incremental manner via network expansion and knowledge distillation. While LST adopts innovative ideas from class-incremental learning~\cite{increment-learning01,increment-learning02,increment-learning03}, the method is overly complex with the maintenance of many sub-parts. Moreover, the method requires numerous stages of knowledge transfer that can lead to catastrophic forgetting and result in an inferior solution~\cite{dong-iccv23}.

Another interesting direction is to leverage external data to augment the training instances in LVIS. The intuition is that while the rare objects may not appear frequently in natural scenes, they can be found in abundance from \emph{object-centric} sources such as ImageNet~\cite{imagenet}. By training on the union of scene-centric \lvis and object-centric ImageNet images, these weakly supervised methods~\cite{mosaicos,detic,richsem} overcome the skewed class distribution by enriching the tail instances with additional whole-image labels. Although appealing, this approach places a hard requirement on having access to a large database of $\sim$14M \emph{labeled} images across $\sim$21K object classes, an assumption not necessarily suitable for many practical settings outside of \lvis.

Consider \Cref{fig::motivation} as an example for aerial imagery; there is no comparable object-centric dataset available to match the semantic concepts of ``track field'', ``roundabout'', or other categories in DOTA~\cite{dota}. Moreover, for an industry application focusing on bespoke concepts beyond general objects, building a new object-centric dataset to supplement the main training dataset can be costly and time-consuming. There is also the open question of how to handle the \emph{genuinely rare} classes (\eg, a rare fish species), which are strictly limited in observation and cannot be collected in more quantity from the open Internet. \emph{Can we still advance long-tailed detection without additional hand-labeling?}

We propose a simple and scalable framework, aptly named \model, to answer this question. We deconstruct the LTD problem into three stages consisting of (i) pre-training on \emph{head} classes, (ii) transfer learning on \emph{tail} classes, and (iii) fine-tuning on a small sample composed of both head and tail classes. Unlike previous research on multi-stage training~\cite{lst,dong-iccv23}, our framework allows for the optional use of \emph{unlabeled images} to further boost LTD. We learn with supplementary unlabeled data in a semi-supervised manner via pseudo-labeling, and therefore do not explicitly rely on any additional instance-level or image-level supervision.

Our work addresses several challenges associated with the head-to-tail model transfer paradigm~\cite{h2t-transfer,MetaModelNet}. For one, we find that the vanilla transfer of data-rich head representations to data-scarce tail classes does not yield sufficient performance improvement because the distribution of head classes is also skewed. Moreover, the na\"ive application of unlabeled data can aggravate the model's inductive bias because the unlabeled samples may follow a similar long-tailed distribution. In such case, the trained model will tend to generate more pseudo labels for the head classes, resulting in a larger degree of pseudo-class imbalance. We overcome these obstacles by incorporating data augmentations specifically designed to mitigate class imbalances in both head and tail training stages. Extensive experiments in \S\ref{sec::main-results} reveal that stronger pre-trained models on head classes, with and without unlabeled images, transfer well to the tail classes, leading to a desirable solution for long-tailed detection.

\medskip
\noindent
\textbf{Main Contributions} ~
\textbf{(1)} We present \model, a general framework for effective supervised and semi-supervised LTD with unlabeled images. \textbf{(2)} The design of \model is straightforward and intuitive, and is compatible with a range of backbones and detectors based on both classical convolutional and modern transformer architectures. \textbf{(3)} When put to the test against competing methods on the challenging \lvis v1 benchmark, \model demonstrates excellent performance and scalability by establishing new state-of-the-art results with compelling margins. We hope \model serves as a strong baseline for future research to tackle realistic long-tailed problems in the community.
\section{Related Work}

\begin{figure*}[t]
\centering
\includegraphics[width=0.93\textwidth]{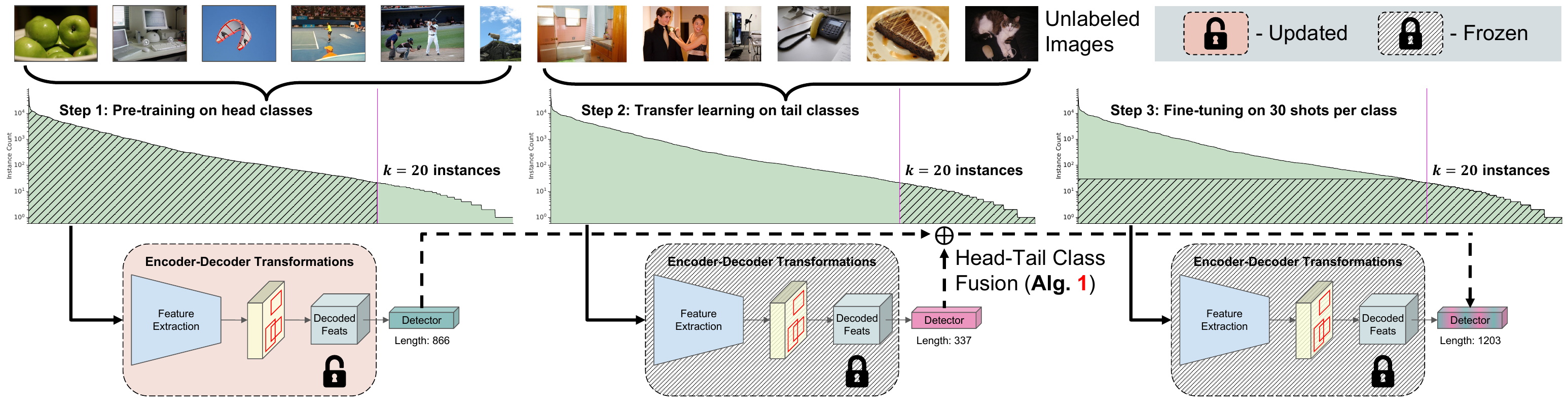}
\caption{Overview of \model. \textbf{Step 1} pre-trains the model and detector on head classes with unlabeled images. \textbf{Step 2} transfers the head representations to tail classes. \textbf{Step 3} fine-tunes the detector on a sample of head and tail exemplar replay. The ``model'' is an abstract block of encoder-decoder transformations based on either a convolutional or transformer network. The model is updated only during pre-training.}
\label{fig::overview}
\end{figure*}

\paragraph{Multi-Stage Learning for LTD}
Multi-stage approaches typically begin with a step on representation learning followed by a transfer learning or fine-tuning routine. The earlier work of LST~\cite{lst} pre-trained a model on a subset of many-shot head classes then transferred its representations to progressively smaller but balanced scarce tail samples via knowledge distillation. A more recent method~\cite{dong-iccv23} pre-trains a model on the entire long-tailed dataset, and opts to fine-tune the resulting model on ``smooth-tail'' data to mitigate the effects of class imbalance. These approaches assume pre-trained representations provide better initialization than random weights for fine-tuning on tail classes.

In \S\ref{sec::pretrained-comparison}, we analyze that assumption to be generally true. We leverage this finding and take a different but related approach to pre-train on the frequent and common head instances. Then we simply copy the head representations and fine-tune them on the tail classes without resorting to knowledge distillation or meta-learning, as was required in prior studies~\cite{MetaModelNet,lst}. Our approach takes three stages of learning compared to the seven stages of LST, which can exacerbate catastrophic forgetting. A novel component to our approach is the introduction of unlabeled images in both pre-training and transfer learning stages for enhanced LTD.

\medskip
\noindent
\textbf{Leveraging Extra Data for LTD} ~
There is an emerging trend of using the abundant weak supervision of ImageNet labels to solve the \lvis problem. Although \lvis is a detection dataset of natural scenes and ImageNet is a classification dataset of object images, they both share 997 overlapping classes between their vocabularies, the intersection of which provides a rich source of $\sim$1.5M extra images to sample additional \lvis instances. To effectively learn from a mixed \lvis-ImageNet dataset with a domain gap, MosaicOS~\cite{mosaicos} leverages mosaic augmentation~\cite{yolov4}, whereas Detic~\cite{detic} and RichSem~\cite{richsem} rely on the CLIP~\cite{clip} classifier to map the semantic concepts between \lvis and ImageNet targets.

As discussed in \S\ref{sec::introduction}, these methods put a strict dependency on a large auxiliary labeled database to augment the main training dataset, which is infeasible for bespoke applications outside of \lvis. Alternatively, we propose a more general and flexible solution to use unlabeled data as a source of auxiliary supervision, which is easy to collect without the burden of human annotations. While our framework is not the first to leverage unlabeled data for LTD, ours is more effective when compared to the competing method of CascadeMatch~\cite{cascadematch}, thanks to our multi-stage training strategy.

\medskip
\noindent
\textbf{Connection to Few-Shot Detection} ~
Long-tailed detection is related to the task of few-shot detection (FSD), the purpose of which is to adapt a base detector (trained on many-shot instances) to learn new concepts from few-shot exemplars. The commonality between the two is obvious---both tasks aim to boost detection on categories with very few training instances. However, LTD has unique challenges that extend beyond FSD. For LTD, the tail classes are authentically rare that follow a Zipf distribution~\cite{zipf} in natural scenes. By contrast, the novel few-shot exemplars in FSD datasets are randomly sampled and are not necessarily rare but can include objects of varying degrees of observational frequency. As such, the multi-stage training methods that work well for FSD~\cite{tfa,ledetection} cannot be directly applied to LTD with the same expected level of effectiveness. We need to devise ways to adapt these methods to the LTD problem to bring improvement over the state of the art.
\section{The \model Framework}
As illustrated in \Cref{fig::motivation}, given a long-tailed dataset \dl with $C$ categories, we split it into two disjoint subsets: \dhead with \chead frequent and common categories appearing in $> M$ images and \dtail with \ctail rare classes appearing in $\le M$ images. Furthermore, we have access to an unlabeled source $\mathcal{D}_\text{u}$ of unknown class distribution. Our goal is to use a combination of labeled and unlabeled images to train a unified model optimized for accuracy on a test set comprising both classes in $C_\text{head} \cup C_\text{tail}$.

Our \model framework consists of three easy steps: (i) representation learning on \dhead, (ii) transfer learning on \dtail, and (iii) fine-tuning on \dk, a reduced dataset composed of $k$ instances per class randomly sampled from \dl. Note that \dhead, \dtail, and \dk are all still imbalanced, but not as severe as the original long-tailed \dl. We leverage \emph{optional} unlabeled images in both Steps 1 and 2 but do not explicitly need them for effective LTD. Indeed, experiments in \S\ref{sec::main-results} show that our fully supervised baselines exhibit excellent performance and scalability even without unlabeled data. A diagram of \model is depicted in \Cref{fig::overview}.

\begin{figure}[t]
\centering
\includegraphics[width=0.7\linewidth]{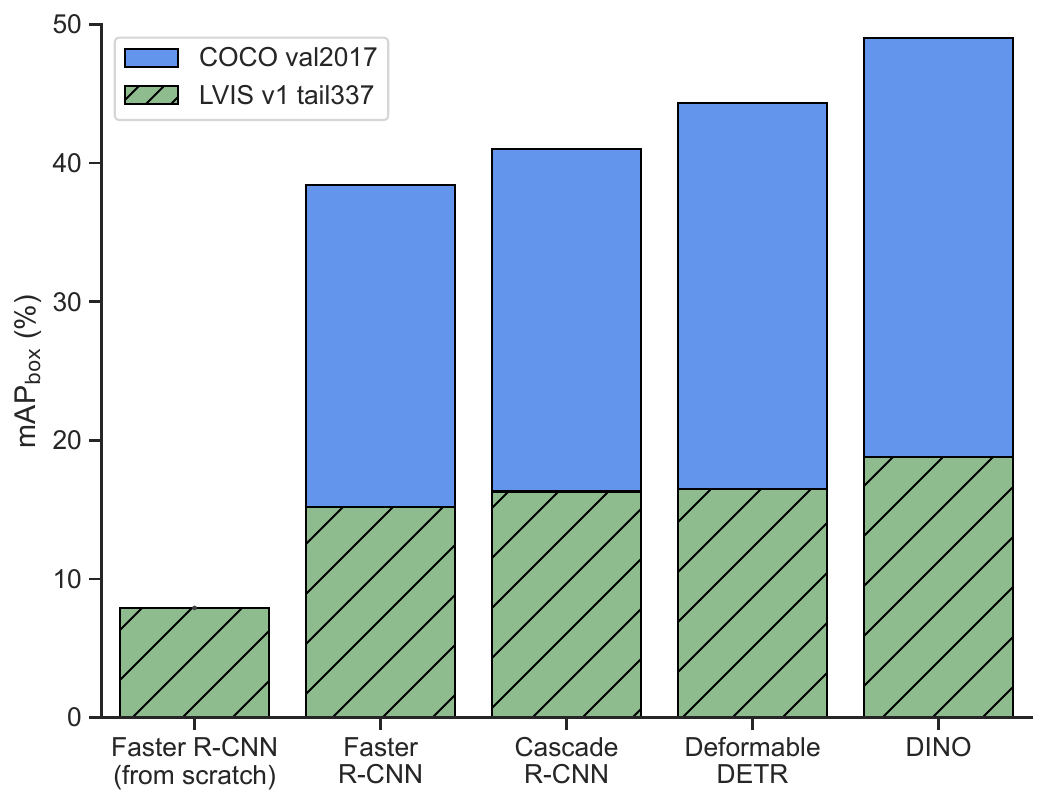}
\caption{Transfer learning from COCO representations (solid bars) helps improve rare class detection on \lvis (hatched green bars).}
\label{fig::pretrained-comparison}
\end{figure}

\subsection{The Devil Is in the \dtail}\label{sec::pretrained-comparison}
The open challenge of the LTD problem remains in learning an effective model on the few exemplars associated with the tail classes. We draw inspiration from existing empirical evidence that few-shot learning can vastly benefit from pre-trained representations~\cite{tfa,ledetection}. With that in mind, we conduct an empirical analysis to verify whether our intuition extends to the \lvis setting. Following conventional \lvis protocol, we partition the dataset into \chead $= 866$ common classes appearing in $M > 10$ images and \ctail $= 337$ rare classes appearing in $M \le 10$ images. We aim to improve the detection performance on \ctail using various commodity detectors pre-trained on the COCO dataset.

\Cref{fig::pretrained-comparison} quantifies the effectiveness of pre-trained representations on \ctail detection. For each model, we chop off the detection head consisting of the bounding box classifier and regressor modules learned on the COCO dataset, re-initialize them with random weights, and perform transfer learning on the \lvis tail classes. We update only the box classifier and regressor while keeping the rest of the architecture frozen, essentially treating the pre-trained model as a fixed detector. Besides the pre-trained networks, we also assess the Faster R-CNN detector~\cite{faster-rcnn} initialized from scratch, except for the pre-trained backbone, as the lower-bound baseline.

\medskip
\noindent
\textbf{Discussion} ~
We observe a clear trend indicating stronger pre-trained representations, as measured by the AP score on COCO, generally lead to improved rare class detection. This result is both intriguing and encouraging since the models were pre-trained on COCO, a dataset of different scope and size than \lvis. Training from scratch brings out the worst performance with half of the accuracy. The implication of this simple experiment is two-fold: (1) we corroborate prior studies by showing low-shot learning can be improved with transferred representations; and (2) our framework opens opportunities to self-supervised, semi-supervised, and multi-modal learning, all of which have demonstrated significantly better performance than supervised pre-training. Motivated by these insights, we propose to learn powerful representations on \dhead and transfer them to \dtail for long-tailed detection. While the past attempts at head-to-tail model transfer~\cite{MetaModelNet,lst,dong-iccv23} could only work by incorporating an extra module for meta-learning or knowledge distillation, we now describe our three-step approach to accomplish this goal without the unnecessary complexities.

\begin{figure*}[t]
  \centering
  \setlength\tabcolsep{1.0pt}
  \begin{tabular}{cccc}
    \makecell{\includegraphics[height=0.201\linewidth]{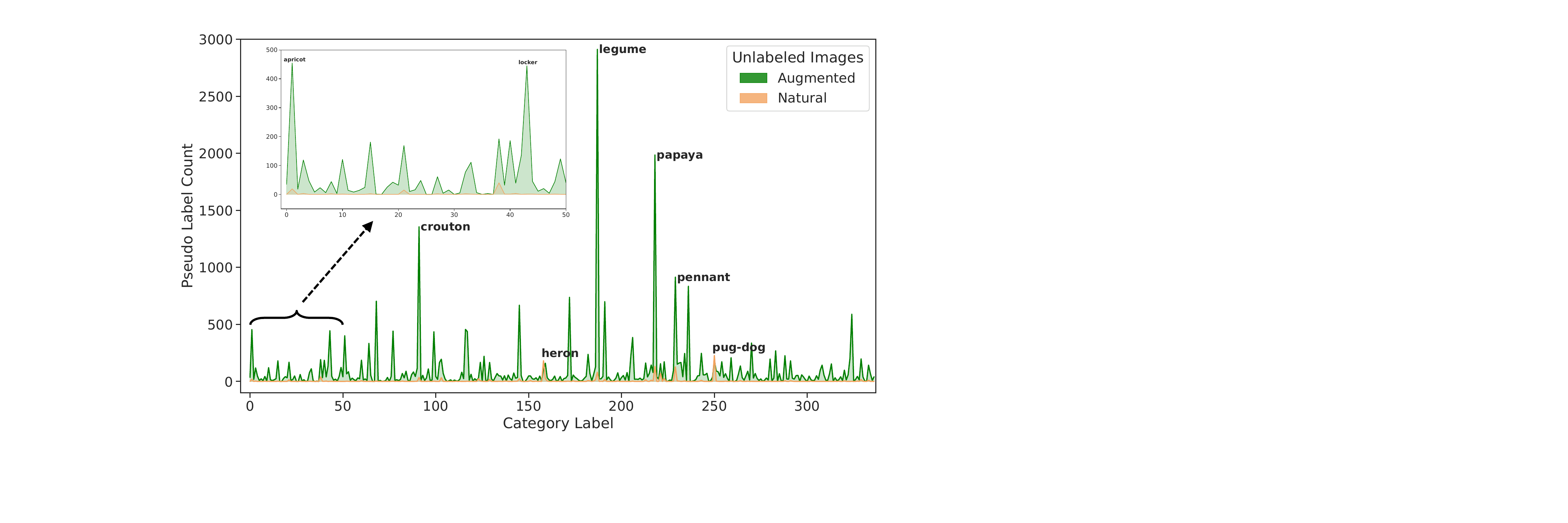}}&
    \makecell{\includegraphics[height=0.195\linewidth]{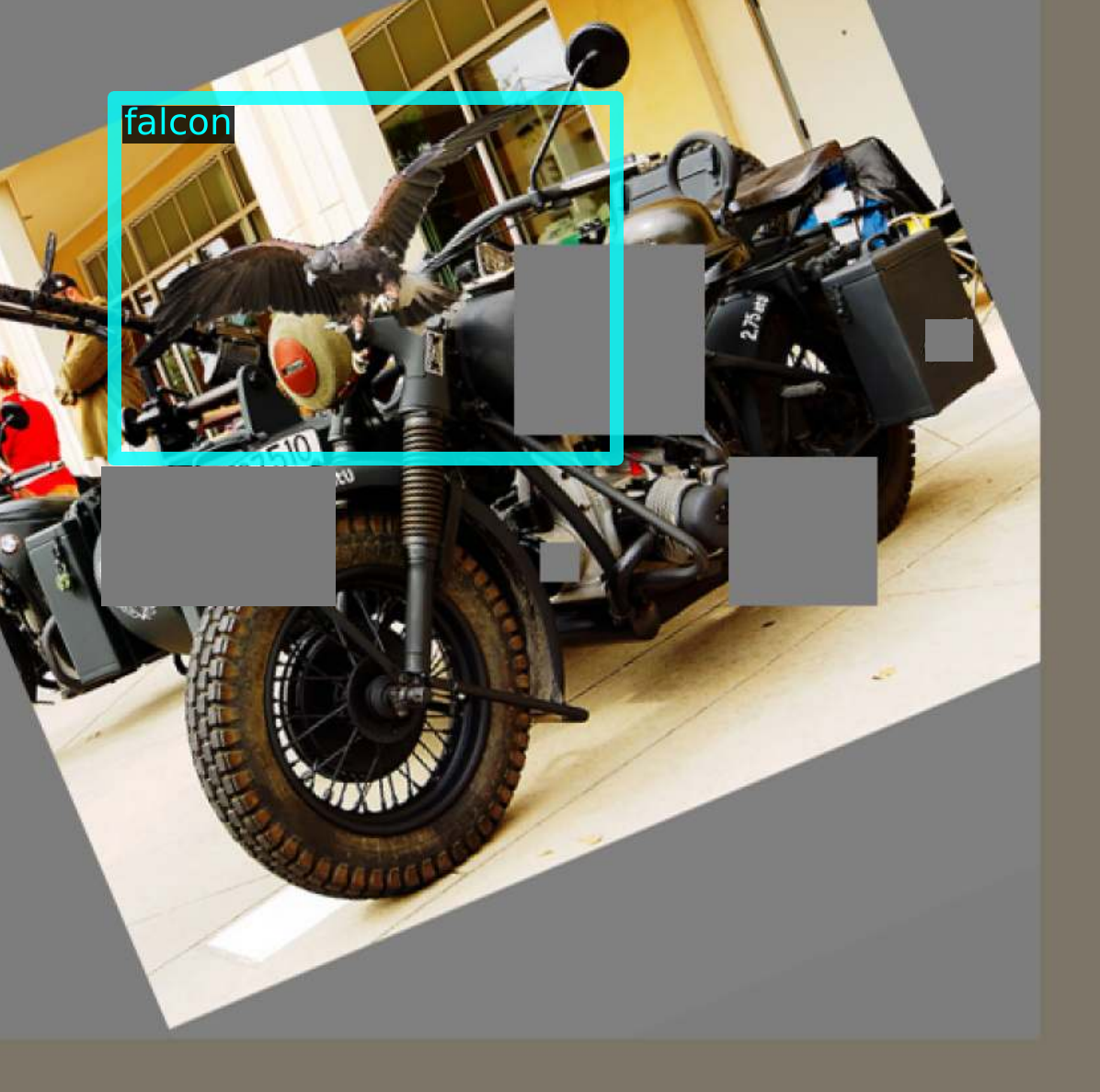}}&
    \makecell{\includegraphics[height=0.195\linewidth]
    {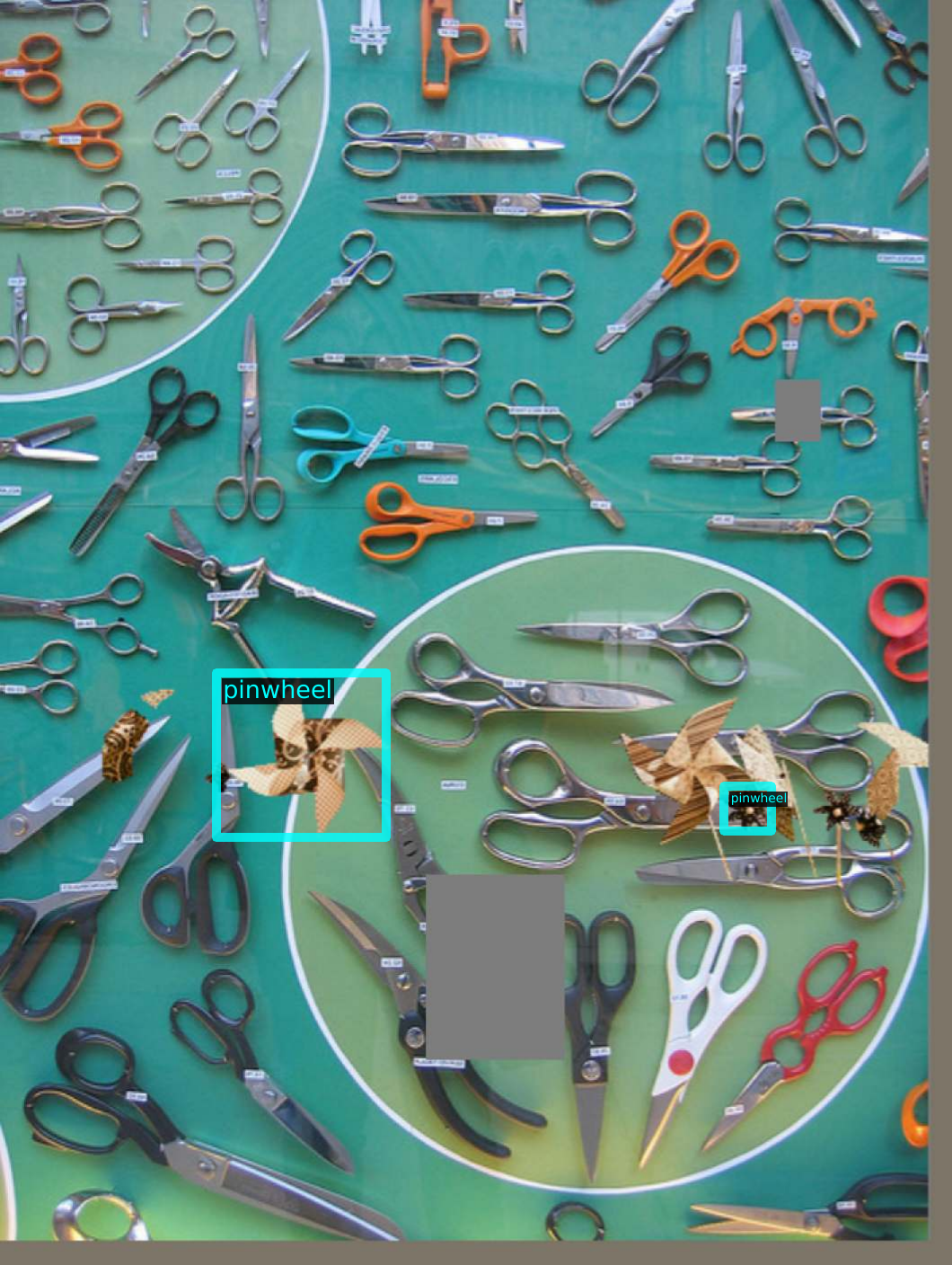}}&
    \makecell{\includegraphics[height=0.195\linewidth]
    {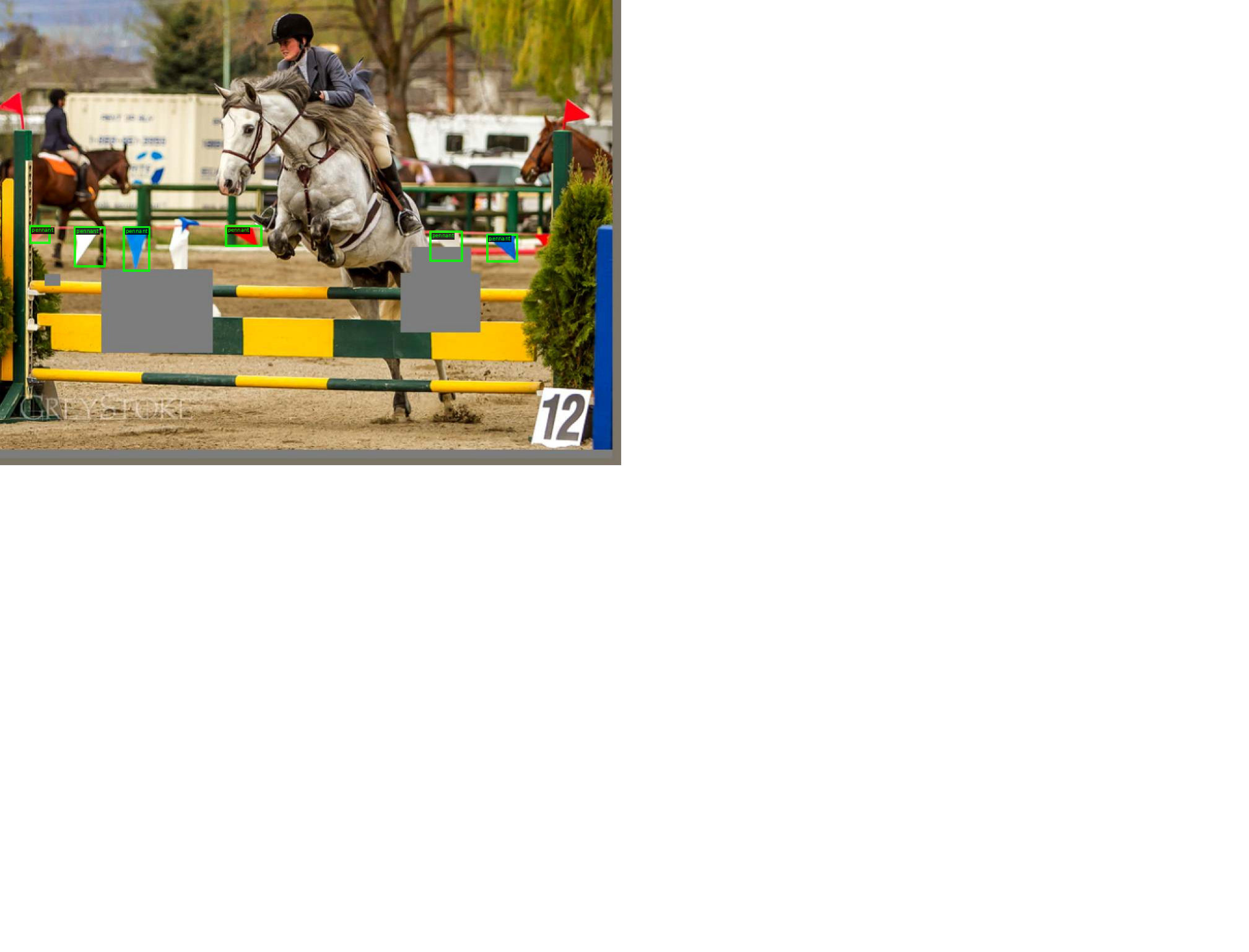}}\\
  \end{tabular}
  \caption{\textbf{Left:} Augmenting unlabeled images with randomly pasted rare instances helps promote pseudo-labeling for effective semi-supervised learning. \textbf{Middle:} The pasted objects (cyan boxes) often come from contrasting environments to create complex \emph{fake} scenes for the student to learn, which in turn can improve robustness and generalization on \emph{natural} scenes (green boxes) shown on the \textbf{Right}.}
  \label{fig::model-analysis}
\end{figure*}
\begin{table}[t]
\centering
\resizebox{0.81\linewidth}{!}{
    \begin{tabular}{lll}
    \toprule
    Augmentation &
    Supervised Training &
    Semi-Supervised Training\\
    \midrule
    \midrule
    Resize
    & short edge $\in [400,1200]$
    & short edge $\in [400,1200]$\\
    Flip
    & horizontal
    & horizontal\\
    SCP with RFS
    & sample threshold $=0.001$
    & sample threshold $=0.001$\\
    AutoContrast
    & \checkmark
    & \checkmark\\
    Equalize
    & \checkmark
    & \checkmark\\
    Solarize
    & \checkmark
    & \checkmark\\
    Color Jittering
    & \checkmark
    & \checkmark\\
    Contrast
    & \checkmark
    & \checkmark\\
    Brightness
    & \checkmark
    & \checkmark\\
    Sharpness
    & \checkmark
    & \checkmark\\
    Posterize
    & \checkmark
    & \checkmark\\
    Translation
    & \multicolumn{1}{l}{\xmark}
    & $(x,y) \in (-0.1,0.1)$\\
    Shearing
    & \multicolumn{1}{l}{\xmark}
    & $(x,y) \in (-30^{\circ}, 30^{\circ})$\\
    Rotation
    & \multicolumn{1}{l}{\xmark}
    & angle $\in (-30^{\circ}, 30^{\circ})$\\
    Cutout
    & \multicolumn{1}{l}{\xmark}
    & $n \in [1,5]$, size $\in [0.0,0.2]$\\
    \bottomrule
    \end{tabular}
}
\caption{Summary of the data augmentations explored in this study to improve supervised and semi-supervised training.}
\label{tab::data-aug}
\end{table}

\subsection{Step 1: Representation Learning on \dhead}\label{sec::step1}
We begin with the supervised setting in which we have training data points $(x_i, y_i) \in$ \dhead, where $x_i$ denotes the $i$th input image and $y_i$ is the $i$th ground truth annotation containing the box label and coordinates. Let $\Psi_\text{det}(\mathcal{D}_\text{head})$ be a learnable detection function training on the image-target pairs to produce the supervised loss $\mathcal{L}_\text{sup}$:
\begin{equation}\label{eq::sup-loss}
    \Psi_\text{det}(\mathcal{D}_\text{head}) \leftarrow \mathcal{L}_\text{sup} =
    \sum L_\text{cls}(h(x_i), y_i) + L_\text{reg}(h(x_i), y_i).
\end{equation}
Here, $h(x_i)$ denotes the forward pass on the input image and $(L_\text{cls}, L_\text{reg})$ are classification (\eg, cross-entropy) and regression (\eg, $L_1$) losses for the detector. We consider commodity convolutional and transformer-based networks for $\Psi_\text{det}$. For the convolutional network, we experiment with Faster R-CNN to compare against previous studies using similar detectors (\ie, Mask R-CNN~\cite{mask-rcnn} and RetinaNet~\cite{retinanet}). For the contemporary transformer-based network, we adopt the improved variants of the Detection Transformer (DETR)~\cite{detr}, namely Deformable DETR~\cite{ddetr} and DINO~\cite{dino}.

As summarized in~\Cref{tab::data-aug}, we leverage well-known data augmentations to train $\Psi_\text{det}$, which include random image resizing, horizontal flipping, and photometric distortion. We also combine Simple Copy-Paste (SCP)~\cite{scp} together with Repeat Factor Sampling (RFS)~\cite{lvis} to combat the class imbalance in \dhead at both the image and instance levels, analogous to prior research on the importance of image and instance resampling for LTD~\cite{simcal,irfs}. Taking a step further to learn even stronger representations on \dhead, we propose to train $\Psi_\text{det}$ in a semi-supervised manner with unlabeled images by way of pseudo-labeling. We explore three successful methods of increasing effectiveness to advance semi-supervised representation learning on \dhead: SoftER Teacher~\cite{ledetection}, MixTeacher~\cite{mixteacher}, and MixPL~\cite{mixpl}.

SoftER Teacher builds upon the end-to-end pseudo-labeling approach of Soft Teacher~\cite{soft-teacher} to include an auxiliary loss for consistency learning on region proposals, and was shown to work particularly well with semi-supervised few-shot detection. MixTeacher introduces a mixed scale feature pyramid to generate more accurate pseudo labels on objects with extreme scale variations, resulting in an overall robust detector. While SoftER Teacher and MixTeacher primarily work with two-stage detectors like Faster R-CNN, MixPL opens the door to semi-supervised learning with single-stage and DETR-based models by integrating mixup~\cite{mixup} and mosaic~\cite{yolov4} augmentations with pseudo labels. All three methods follow the student-teacher semi-supervised training paradigm~\cite{mean-teacher}, in which the teacher is an exponential moving average of the student. In the semi-supervised setting, the models learn from a joint dataset of labeled \dhead and unlabeled $\mathcal{D}_\text{u}$ images via the following compound objective:
\vspace{-5pt}
\begin{equation}\label{eq::semi-sup-loss}
    \Psi_\text{semi-det}(\mathcal{D}_\text{head},\mathcal{D}_\text{u}) \leftarrow \mathcal{L} =
    \mathcal{L}_\text{sup} + \alpha \mathcal{L}_\text{pseudo},
\end{equation}
where $\alpha > 0$ controls the contribution of the pseudo-label loss derived from unlabeled data. The functional form of $\mathcal{L}_\text{pseudo}$ is the same as $\mathcal{L}_\text{sup}$ in \Cref{eq::sup-loss}, except the ground-truth targets $y$ are replaced by pseudo targets $\hat{y}$ predicted by the teacher model during self-training.

\begin{table*}[t]
\centering
\resizebox{0.98\textwidth}{!}{
    \begin{tabular}{lcllrrrrr}
        \toprule
        Method &
        Extra External Data &
        Base Detector &
        Backbone &
        GPU Hrs &
        m\apbox &
        \apr &
        \apc &
        \apf\\
        \midrule
        \midrule
        Seesaw Loss [CVPR'21]~\cite{seesaw} &
        \multicolumn{1}{c}{\multirow{6}{*}{N/A}} &
        Faster R-CNN &
        R101-FPN &
        -- &
        27.8 &
        18.7 &
        27.0 &
        32.8\\
        NorCal [NeurIPS'21]~\cite{norcal} &
        &
        Mask R-CNN &
        R101-FPN &
        -- &
        28.1 &
        20.8 &
        26.5 &
        30.9\\
        AHRL [CVPR'22]~\cite{ahrl} &
        &
        Mask R-CNN &
        R101-FPN &
        -- &
        28.7 &
        19.3 &
        27.6 &
        31.4\\
        EFL [CVPR'22]~\cite{efl} &
        &
        RetinaNet &
        R101-FPN &
        -- &
        29.2 &
        23.5 &
        27.4 &
        33.8\\
        \cellcolor{Gray}\model Supervised &
        \cellcolor{Gray}&
        \cellcolor{Gray}Faster R-CNN &
        \cellcolor{Gray}R101-FPN &
        \cellcolor{Gray}83 &
        \cellcolor{Gray}\textbf{31.7} &
        \cellcolor{Gray}\textbf{24.3} &
        \cellcolor{Gray}\textbf{32.1} &
        \cellcolor{Gray}\textbf{34.6}\\
        \cellcolor{Gray}\model Supervised &
        \cellcolor{Gray}&
        \cellcolor{Gray}Deformable DETR &
        \cellcolor{Gray}R101 &
        \cellcolor{Gray}234 &
        \cellcolor{Gray}\textbf{37.0} &
        \cellcolor{Gray}\textbf{31.9} &
        \cellcolor{Gray}\textbf{36.6} &
        \cellcolor{Gray}\textbf{39.6}\\
        \midrule
        Dong et al. [ICCV'23]~\cite{dong-iccv23} &
        \multicolumn{1}{c}{\multirow{5}{*}{N/A}} &
        Deformable DETR &
        R50 &
        -- &
        28.7 &
        21.8 &
        28.4 &
        32.0\\
        Detic [ECCV'22]~\cite{detic} &
        &
        Deformable DETR &
        R50 &
        -- &
        31.7 &
        21.4 &
        30.7 &
        37.5\\
        RichSem [NeurIPS'23]~\cite{richsem} &
        &
        DINO &
        R50 &
        -- &
        \textbf{35.1} &
        26.0 &
        32.6 &
        \textbf{41.8}\\
        \cellcolor{Gray}\model Supervised &
        \cellcolor{Gray}&
        \cellcolor{Gray}Faster R-CNN &
        \cellcolor{Gray}R50-FPN &
        \cellcolor{Gray}81 &
        \cellcolor{Gray}29.0 &
        \cellcolor{Gray}20.9 &
        \cellcolor{Gray}29.7 &
        \cellcolor{Gray}31.9\\
        \cellcolor{Gray}\model Supervised &
        \cellcolor{Gray}&
        \cellcolor{Gray}Deformable DETR &
        \cellcolor{Gray}R50 &
        \cellcolor{Gray}215 &
        \cellcolor{Gray}35.0 &
        \cellcolor{Gray}\textbf{32.0} &
        \cellcolor{Gray}\textbf{34.0} &
        \cellcolor{Gray}37.5\\
        \midrule
        RichSem [NeurIPS'23]~\cite{richsem} &
        \multicolumn{1}{c}{\multirow{7}{*}{N/A}} &
        DINO &
        Swin-T &
        -- &
        38.8 &
        30.8 &
        36.4 &
        45.0\\
        \cellcolor{Gray}\model Supervised &
        \cellcolor{Gray}&
        \cellcolor{Gray}DINO &
        \cellcolor{Gray}Swin-T &
        \cellcolor{Gray}310 &
        \cellcolor{Gray}\textbf{41.1} &
        \cellcolor{Gray}\textbf{33.6} &
        \cellcolor{Gray}\textbf{40.1} &
        \cellcolor{Gray}\textbf{45.4}\\
        DiffusionDet [ICCV'23]~\cite{diffusiondet} &
        &
        4 @ 300 &
        Swin-B &
        -- &
        42.0 &
        34.8 &
        40.9 &
        46.4\\
        RichSem [NeurIPS'23]~\cite{richsem} &
        &
        DINO &
        Swin-B &
        -- &
        46.4 &
        38.5 &
        45.1 &
        \textbf{51.3}\\
        \cellcolor{Gray}\model Supervised &
        \cellcolor{Gray}&
        \cellcolor{Gray}DINO &
        \cellcolor{Gray}Swin-B &
        \cellcolor{Gray}414 &
        \cellcolor{Gray}\textbf{47.2} &
        \cellcolor{Gray}\textbf{42.7} &
        \cellcolor{Gray}\textbf{46.7} &
        \cellcolor{Gray}49.9\\
        RichSem [NeurIPS'23]~\cite{richsem} &
        &
        DINO &
        Swin-L &
        -- &
        49.7 &
        \textbf{42.8} &
        49.2 &
        \textbf{53.4}\\
        \cellcolor{Gray}\model Supervised &
        \cellcolor{Gray}&
        \cellcolor{Gray}DINO &
        \cellcolor{Gray}Swin-L &
        \cellcolor{Gray}460 &
        \cellcolor{Gray}\textbf{49.8} &
        \cellcolor{Gray}42.4 &
        \cellcolor{Gray}\textbf{50.4} &
        \cellcolor{Gray}52.4\\
        \midrule
        MosaicOS [ICCV'21]~\cite{mosaicos} &
        ImageNet-1K Labels &
        Faster R-CNN &
        R50-FPN &
        -- &
        23.9 &
        15.5 &
        22.4 &
        29.3\\
        RichSem [NeurIPS'23]~\cite{richsem} &
        ImageNet-21K Labels &
        Faster R-CNN $+$ CLIP &
        R50-FPN &
        -- &
        30.6 &
        \textbf{27.6} &
        29.7 &
        32.9\\
        \cellcolor{Gray}\model SoftER Teacher~\cite{ledetection} &
        \cellcolor{Gray}COCO-unlabeled2017 &
        \cellcolor{Gray}Faster R-CNN &
        \cellcolor{Gray}R50-FPN &
        \cellcolor{Gray}392 &
        \cellcolor{Gray}30.3 &
        \cellcolor{Gray}23.3 &
        \cellcolor{Gray}30.3 &
        \cellcolor{Gray}33.3\\
        \cellcolor{Gray}\model MixTeacher~\cite{mixteacher} &
        \cellcolor{Gray}COCO-unlabeled2017 &
        \cellcolor{Gray}Faster R-CNN &
        \cellcolor{Gray}R50-FPN &
        \cellcolor{Gray}434 &
        \cellcolor{Gray}\textbf{31.8} &
        \cellcolor{Gray}23.4 &
        \cellcolor{Gray}\textbf{32.1} &
        \cellcolor{Gray}\textbf{35.1}\\
        \midrule
        Detic [ECCV'22]~\cite{detic} &
        ImageNet-21K Labels &
        DeformDETR $+$ CLIP &
        R50 &
        -- &
        32.5 &
        26.2 &
        31.3 &
        36.6\\
        RichSem [NeurIPS'23]~\cite{richsem} &
        ImageNet-21K Labels &
        DINO $+$ CLIP &
        R50 &
        -- &
        37.1 &
        29.9 &
        35.6 &
        42.0\\
        \cellcolor{Gray}\model MixPL~\cite{mixpl} &
        \cellcolor{Gray}Objects365-unlabeled &
        \cellcolor{Gray}DINO &
        \cellcolor{Gray}R50 &
        \cellcolor{Gray}447 &
        \cellcolor{Gray}39.1 &
        \cellcolor{Gray}31.5 &
        \cellcolor{Gray}\textbf{38.5} &
        \cellcolor{Gray}43.1\\
        \cellcolor{Gray}\model MixPL~\cite{mixpl} &
        \cellcolor{Gray}COCO-unlabeled2017 &
        \cellcolor{Gray}DINO &
        \cellcolor{Gray}R50 &
        \cellcolor{Gray}446 &
        \cellcolor{Gray}\textbf{39.4} &
        \cellcolor{Gray}\textbf{32.6} &
        \cellcolor{Gray}\textbf{38.5} &
        \cellcolor{Gray}\textbf{43.6}\\
        \midrule
        RichSem [NeurIPS'23]~\cite{richsem} &
        ImageNet-21K Labels &
        DINO $+$ CLIP &
        Swin-T &
        -- &
        41.6 &
        \textbf{37.3} &
        39.7 &
        45.5\\
        \cellcolor{Gray}\model MixPL~\cite{mixpl} &
        \cellcolor{Gray}COCO-unlabeled2017 &
        \cellcolor{Gray}DINO &
        \cellcolor{Gray}Swin-T &
        \cellcolor{Gray}482 &
        \cellcolor{Gray}\textbf{42.5} &
        \cellcolor{Gray}35.7 &
        \cellcolor{Gray}\textbf{42.4} &
        \cellcolor{Gray}\textbf{45.6}\\
        RichSem [NeurIPS'23]~\cite{richsem} &
        ImageNet-21K Labels &
        DINO $+$ CLIP &
        Swin-B &
        -- &
        48.2 &
        \textbf{46.5} &
        46.5 &
        51.0\\
        \cellcolor{Gray}\model MixPL~\cite{mixpl} &
        \cellcolor{Gray}COCO-unlabeled2017 &
        \cellcolor{Gray}DINO &
        \cellcolor{Gray}Swin-B &
        \cellcolor{Gray}794 &
        \cellcolor{Gray}\textbf{49.0} &
        \cellcolor{Gray}43.4 &
        \cellcolor{Gray}\textbf{49.0} &
        \cellcolor{Gray}\textbf{51.5}\\
        RichSem [NeurIPS'23]~\cite{richsem} &
        ImageNet-21K Labels &
        DINO $+$ CLIP &
        Swin-L &
        -- &
        \textbf{52.0} &
        \textbf{50.2} &
        51.5 &
        \textbf{53.3}\\
        \cellcolor{Gray}\model MixPL~\cite{mixpl} &
        \cellcolor{Gray}COCO-unlabeled2017 &
        \cellcolor{Gray}DINO &
        \cellcolor{Gray}Swin-L &
        \cellcolor{Gray}1168 &
        \cellcolor{Gray}51.5 &
        \cellcolor{Gray}45.0 &
        \cellcolor{Gray}\textbf{52.4} &
        \cellcolor{Gray}\textbf{53.3}\\
        \bottomrule
    \end{tabular}
}
\vspace{-3pt}
\captionof{table}{Main results on \lvis v1 validation. GPU hours denote the wall clock time to train for a total of 640K iterations and are a proxy measure of model complexity. The ResNet and Swin backbones were pre-trained on ImageNet-1K and ImageNet-22K, respectively. The results of Seesaw Loss and Detic are borrowed from EFL and RichSem, respectively. \colorbox{Gray}{Shaded rows} indicate our implemented models.}
\label{tab::main-results}
\end{table*}

\subsection{Step 2: Transfer Learning on \dtail}\label{sec::step2}
We instantiate the tail models for transfer learning by copying the parameters from the pre-trained head models. Let $\Psi'_\text{det}(\mathcal{D}_\text{tail}) \xleftarrow{} \Psi_\text{det}(\mathcal{D}_\text{head})$ be the supervised tail model and $\Psi'_\text{semi-det}(\mathcal{D}_\text{tail}, \mathcal{D}_\text{u}) \xleftarrow{} \Psi_\text{semi-det}$ be the semi-supervised counterpart. We train the tail models the same way per \Cref{eq::sup-loss,eq::semi-sup-loss}, except we update only the classifier and regressor modules, to adapt them to tail classes, while freezing the rest of the networks. The intuition is that pre-trained representations serve as a bootstrapped initializer to learn an accurate tail model, according to our analysis in \S\ref{sec::pretrained-comparison}.

Recall that, unlike common head objects, the tail classes are intrinsically rare so one cannot expect to find abundant occurrences of them in either labeled or unlabeled source. This raises a hurdle for when we wish to train $\Psi'_\text{semi-det}$ with unlabeled images: there are very few instances of the rare classes in the unlabeled scenes for the teacher model to propose reliable pseudo targets. We sidestep this hurdle by copying and pasting a random subset of rare instances from the labeled training set to the unlabeled images---a new procedure unique to this work. At each training iteration, the teacher model is guaranteed to see an augmented view of sampled rare objects, amid diverse background scenes, which promotes pseudo-label supervision for the student model. \Cref{fig::model-analysis} illustrates the impact of this technique on pseudo-labeling along with some examples of the augmented unlabeled images, which are subjected to strong photometric and geometric perturbations with cutout~\cite{cutout,random-erase}. Although the procedure inevitably leads to redundancy and overfitting, our ablation experiments in \S\ref{sec::ablation} reveal that it is surprisingly helpful in adapting head representations to the tail models.

\begin{algorithm}[t]
\caption{\small PyTorch Pseudocode for Head-Tail Class Fusion.}
\label{algorithm}
\definecolor{codeblue}{rgb}{0.25,0.5,0.5}
\definecolor{codekw}{rgb}{0.85, 0.18, 0.50}
\lstset{
  backgroundcolor=\color{white},
  basicstyle=\fontsize{7.5pt}{7.5pt}\ttfamily\selectfont,
  columns=fullflexible,
  breaklines=true,
  captionpos=b,
  commentstyle=\color{codeblue},
  keywordstyle=\color{codekw},
}
\begin{lstlisting}[language=python]
# HEAD_IDS : sorted list of head IDs, length 866
# TAIL_IDS : sorted list of tail IDs, length 337
# head_ckpt: model checkpoint on head classes
# tail_ckpt: model checkpoint on tail classes

ALL_IDS = sorted(HEAD_IDS + TAIL_IDS) # length 1203
ID2LABEL = {
    ID: label for label, ID in enumerate(ALL_IDS)
} # mapping from category ID to integer label
head_det = head_ckpt["state_dict"]["detector"]
tail_det = tail_ckpt["state_dict"]["detector"]
fused_det = torch.randn(len(ALL_IDS))

for label, ID in enumerate(HEAD_IDS):
    fused_det[ID2LABEL[ID]] = head_det[label]
for label, ID in enumerate(TAIL_IDS):
    fused_det[ID2LABEL[ID]] = tail_det[label]

head_ckpt["state_dict"]["detector"] = fused_det
torch.save(head_ckpt, save_filename) # to fine-tune
\end{lstlisting}
\end{algorithm}

\subsection{Step 3: Fine-Tuning on \dk}\label{sec::step3}
At this stage, we have two separate models with a shared representation: one optimized on head classes and the other on tail classes. We wish to unify the two models into one for efficient single-pass inference on test samples containing both head and tail classes. \Cref{algorithm} gives the pseudocode for the merging scheme as referenced in \Cref{fig::overview}.

Note that we merge parameters at the detector module and reuse the pre-trained head representations for the rest of the network. We fine-tune the unified detector on \dk composed of $k$ instances, or shots, per class sampled from the training set. We form \dk to include both head and tail categories for exemplar replay. We update only the box classifier and regressor with a reduced learning rate to slowly adapt them to the tail classes, while preserving the pre-trained accuracy to avoid catastrophic forgetting on the head classes.
\section{Empirical Study}

\begin{figure*}[t]
  \centering
  \setlength\tabcolsep{1.0pt}
  \begin{tabular}{cccc}
    \makecell{\includegraphics[height=0.21\linewidth]{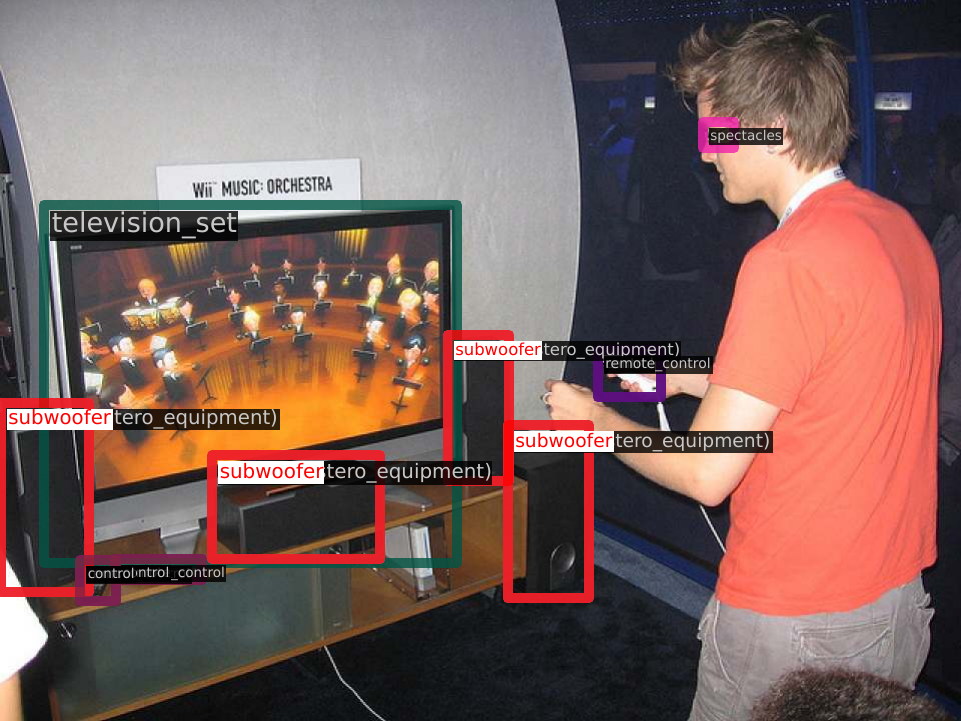}}&
    \makecell{\includegraphics[height=0.21\linewidth]{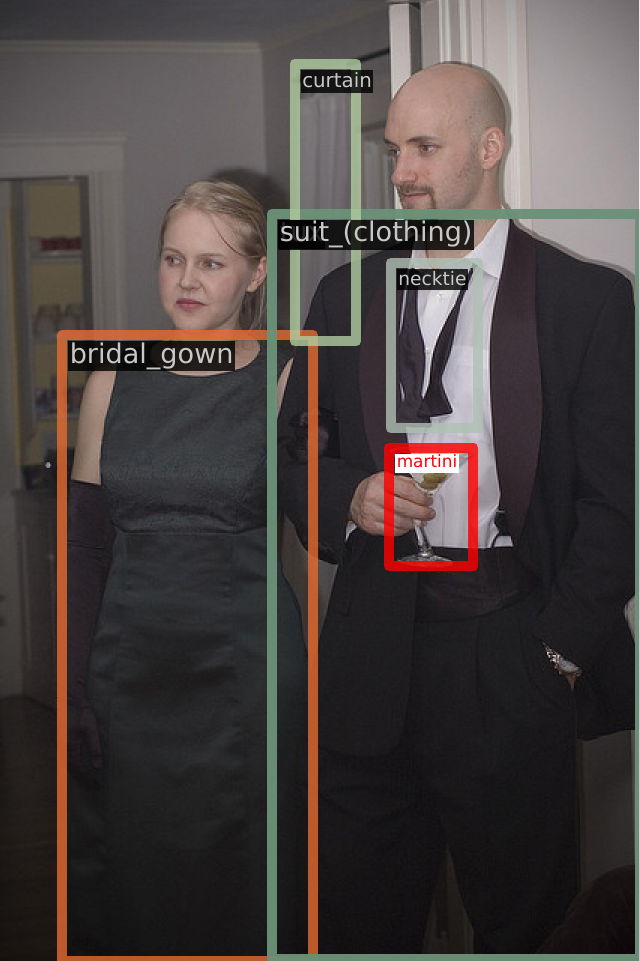}}&
    \makecell{\includegraphics[height=0.21\linewidth]{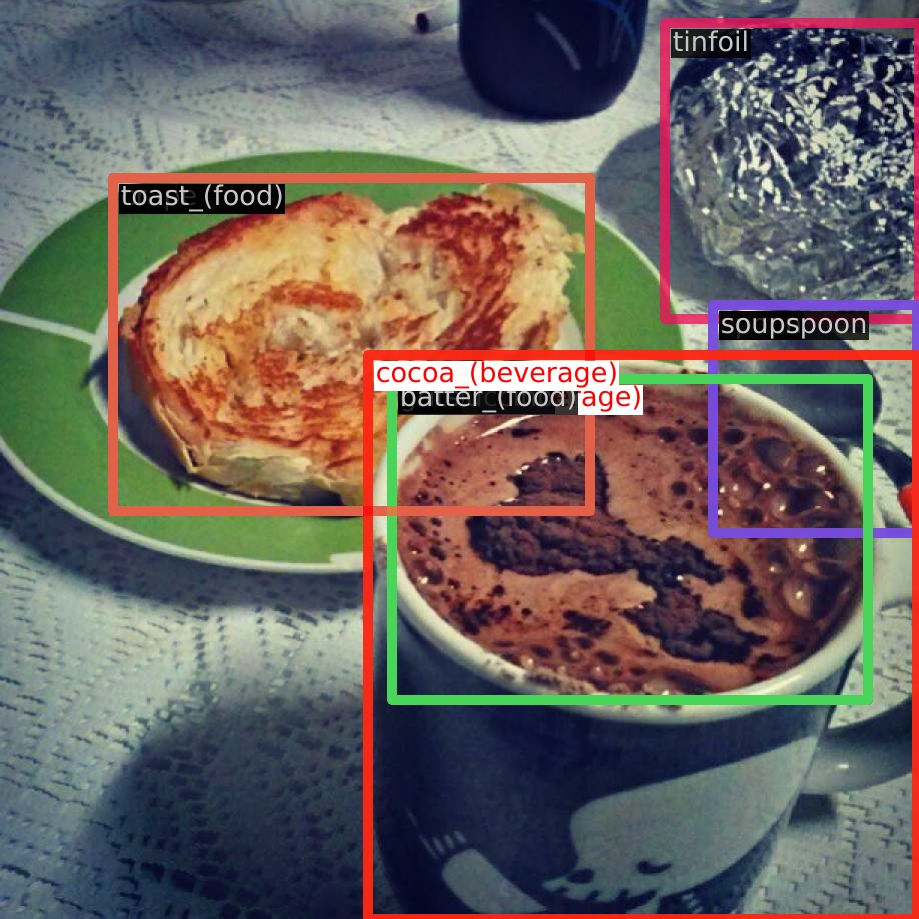}}&
    \makecell{\includegraphics[height=0.21\linewidth]{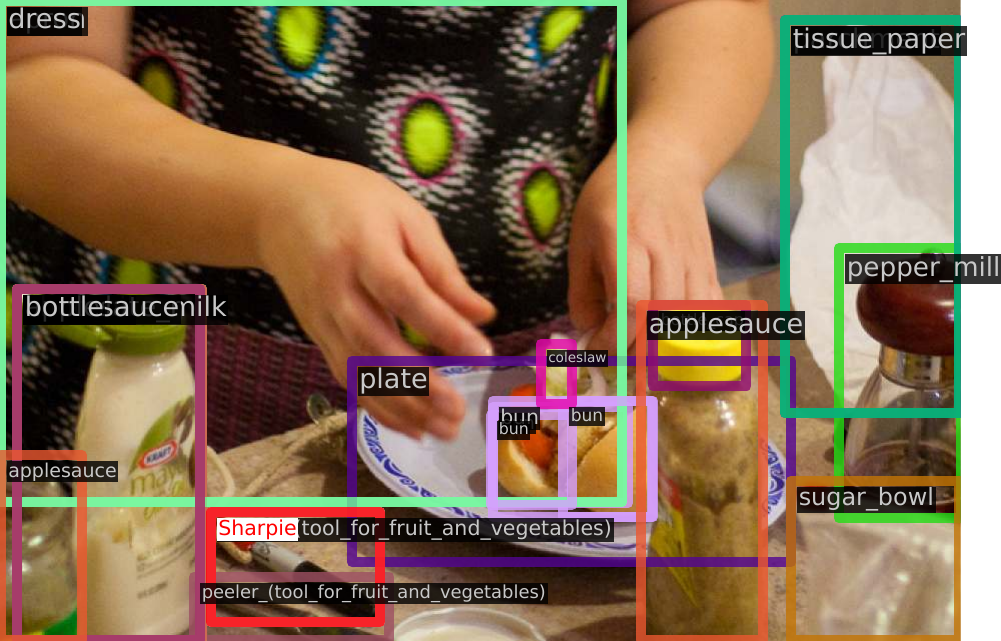}}\\
    \footnotesize $k=1$ shot & \footnotesize $k=3$ shots & \footnotesize $k=4$ shots & \footnotesize $k=5$ shots
  \end{tabular}
  \vspace{-3pt}
  \caption{\model detections on \lvis v1 validation images. We highlight visualizations containing truly rare $k$-shot exemplars from the training set---drawn with red and white boxes. \model does well in the extreme low-shot regime using as few as a single training instance.}
  \label{fig::qualitative-vis}
\end{figure*}

\subsection{Evaluation Protocol}
We benchmark our approach on the challenging \lvis v1 dataset, which has $100170$ training and $19809$ validation images over $1203$ classes. We compute the detection metric m\apbox for all categories, without test-time augmentation, along with \apr, \apc, and \apf for the rare, common, and frequent classes. We sample \dk three times with different random seeds and report on the averaged metrics to capture statistical variability. Following prior studies, we focus on the gains of m\apbox and \apr in our comparative analysis.

\subsection{Implementation Details}
We implement our models using PyTorch~\cite{pytorch} and MMDetection~\cite{mmdetection}, and train on 8$\times$ A6000 GPUs. We pre-train in Step 1 for up to 540K iterations, perform transfer learning in Step 2 for 20K iterations, and fine-tune for 80K iterations. See our open-source code for the full reproducible details.

\medskip
\noindent
\textbf{High-Quality Supervised Baseline} ~
We construct a strong supervised baseline by combining our training recipe with diverse data augmentations. We explore various ResNet~\cite{resnet} and Swin~\cite{swin} backbones, along with FPN~\cite{fpn}, for feature extraction. The supervised baseline is important to our framework because it serves as the basis for the teacher model to propose reliable pseudo targets for semi-supervised learning.

\medskip
\noindent
\textbf{Semi-Supervised LTD} ~
We leverage SoftER Teacher, MixTeacher, and MixPL for semi-supervised LTD, and inherit all hyper-parameters originally tuned on the COCO dataset without changes. We harness $\sim$123K COCO-unlabeled2017 images to improve both representation and transfer learning in Steps 1 and 2. We also experiment with Objects365~\cite{objects365} to further validate our approach on another related domain with $\sim$1.7M unlabeled images in the wild by removing all label information from the training set.

\begin{figure}[t]
\centering
\includegraphics[width=0.99\linewidth]{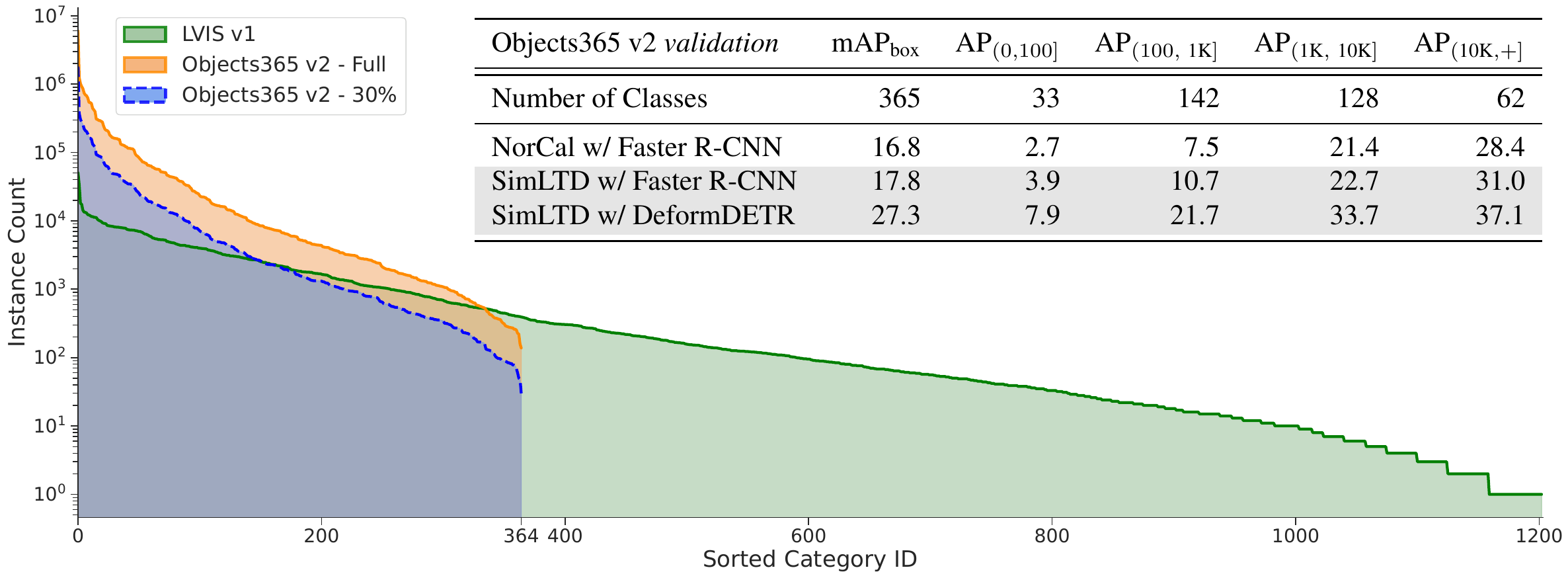}
\vspace{-1.75pt}
\caption{Evaluation on Objects365 binned by the count of training images per class group. All models use the ResNet-50 backbone.}
\label{fig::objects365-results}
\end{figure}

\subsection{Main Results}\label{sec::main-results}
\Cref{tab::main-results} reports on the effectiveness of our approach against existing methods representing the state of the art on \lvis. Our \model supervised baseline with Faster R-CNN outperforms all methods using related detectors. The gains are convincing, with margins up to $+3.9$ \apbox and $+5.6$ \apr. We observe a similar trend when comparing our supervised baseline using DETR-based models. \model demonstrates compelling performance and scalability across a multitude of backbones and detectors without the need for extra data.

For the methods requiring external data, our semi-supervised models also deliver impressive performance. When equipped with MixTeacher and Faster R-CNN, \model exceeds the competition by up to $+7.9$ \apbox while being competitive on \apr. Furthermore, \model scales well by coupling with MixPL and transformer-based models to achieve new state-of-the-art results from harnessing only unlabeled images. \model works equally well with both COCO and Objects365 unlabeled images, signifying that our model can extract meaningful pseudo-label supervision from a large uncurated database with a distribution different from the training dataset. \Cref{fig::qualitative-vis} visualizes example \model detections on select \lvis validation images.

Ancillary to the main \lvis results, we also evaluate on Objects365 to showcase the generality of our framework. Following NorCal~\cite{norcal}, we sample 30\% of the training set and split it into \chead $= 332$ classes appearing in $M > 100$ images and \ctail $= 33$ classes appearing in $M \le 100$ images. \model outperforms the existing baseline across the board in \Cref{fig::objects365-results}. We fine-tune \model with $k=30000$ shots, a parameter that changes by dataset. \Cref{sec::ablation} gives a detailed ablation analysis regarding the impact on AP from varying the number of shots for fine-tuning with \dk.

\begin{table}
\centering
\resizebox{\linewidth}{!}{
    \begin{tabular}{lllrrrr}
    \toprule
    Method &
    Base Detector &
    Backbone &
    AP$_\text{box}^\text{Fixed}$ &
    AP$_\text{r}^\text{Fixed}$ &
    AP$_\text{c}^\text{Fixed}$ &
    AP$_\text{f}^\text{Fixed}$\\
    \midrule
    \midrule
    CascadeMatch (Sup) &
    Cascade R-CNN &
    \multirow{3}{*}{R101-FPN} &
    27.1 &
    20.3 &
    26.1 &
    31.1\\
    FASA (Supervised) &
    Mask R-CNN &
    &
    28.2 &
    22.0 &
    28.3 &
    30.9\\
    \cellcolor{Gray}\model Supervised &
    \cellcolor{Gray}Faster R-CNN &
    \cellcolor{Gray}&
    \cellcolor{Gray}\textbf{32.7} &
    \cellcolor{Gray}\textbf{24.9} &
    \cellcolor{Gray}\textbf{32.9} &
    \cellcolor{Gray}\textbf{35.9}\\
    \midrule
    CascadeMatch &
    Cascade R-CNN &
    &
    30.5 &
    23.1 &
    29.7 &
    34.7\\
    \cellcolor{Gray}\model SoftER Teacher &
    \cellcolor{Gray}Faster R-CNN &
    \cellcolor{Gray}R50-FPN &
    \cellcolor{Gray}31.3 &
    \cellcolor{Gray}24.1 &
    \cellcolor{Gray}31.1 &
    \cellcolor{Gray}34.6\\
    \cellcolor{Gray}\model MixTeacher &
    \cellcolor{Gray}Faster R-CNN &
    \cellcolor{Gray}&
    \cellcolor{Gray}\textbf{32.8} &
    \cellcolor{Gray}\textbf{24.5} &
    \cellcolor{Gray}\textbf{32.9} &
    \cellcolor{Gray}\textbf{36.4}\\
    \midrule
    CascadeMatch &
    Cascade R-CNN &
    &
    32.9 &
    \textbf{26.5} &
    31.8 &
    36.8\\
    \cellcolor{Gray}\model SoftER Teacher &
    \cellcolor{Gray}Faster R-CNN &
    \cellcolor{Gray}R101-FPN &
    \cellcolor{Gray}33.0 &
    \cellcolor{Gray}26.1 &
    \cellcolor{Gray}32.6 &
    \cellcolor{Gray}36.4\\
    \cellcolor{Gray}\model MixTeacher &
    \cellcolor{Gray}Faster R-CNN &
    \cellcolor{Gray}&
    \cellcolor{Gray}\textbf{34.4} &
    \cellcolor{Gray}26.1 &
    \cellcolor{Gray}\textbf{34.2} &
    \cellcolor{Gray}\textbf{38.2}\\
    \bottomrule
    \end{tabular}
}
\vspace{2pt}
\captionof{table}{Performance comparison between \model, FASA~\cite{fasa}, and CascadeMatch~\cite{cascadematch} using the alternative AP$^\text{Fixed}$ metric~\cite{fixed-ap}.}
\label{tab::results-apfixed}
\end{table}

\subsection{Comparison to the State of the Art}

\paragraph{\model \vs CascadeMatch~\cite{cascadematch}}
Although both \model and CascadeMatch leverage COCO-unlabeled2017 for semi-supervised LTD, there are major differences between the two. First, CascadeMatch is trained end-to-end in a single stage, whereas we take the decoupled approach. Second, CascadeMatch adopts the stronger Cascade R-CNN~\cite{cascade-rcnn} compared to Faster R-CNN in \model. CascadeMatch follows the AP$^\text{Fixed}$ protocol~\cite{fixed-ap}, which replaces the standard maximum 300 detections per image by a cap of 10K detections per class from the entire validation set. Despite the disadvantage of a simpler model, \Cref{tab::results-apfixed} shows that \model outperforms CascadeMatch by notable margins in almost every measure. These superior results lend further support to the merit of our multi-stage training strategy.

\medskip
\noindent
\textbf{\model \vs Dong et al.~\cite{dong-iccv23} on Multi-Stage Learning} ~
We compare our multi-stage training strategy to that of Dong et al., which also utilizes a three-step procedure of pre-training followed by fine-tuning and knowledge distillation. We focus our analysis on their powerful Deformable DETR model, which yields analogous results to our simpler Faster R-CNN model. When we train with the same capable Deformable DETR architecture, \Cref{tab::main-results} shows that \model exceeds their model by outsized margins of $+6.3$ \apbox and $+10.2$ \apr. These remarkable gains are directly attributed to our multi-stage learning approach, which is carefully designed to optimize for accuracy on both head and tail classes.

\medskip
\noindent
\textbf{\model \vs RichSem~\cite{richsem} on Using Extra Data} ~
Recall that RichSem relies on the CLIP classifier (pre-trained on $\sim$400M image-caption pairs) and image-supervision from an additional $\sim$1.5M images to produce the state-of-the-art results reported in \Cref{tab::main-results}. However, such strict dependencies are impractical when the method is applied to a bespoke dataset outside of generic objects. By contrast, our \model leverages unlabeled images, without resorting to either CLIP or auxiliary ImageNet labels, to deliver better results than RichSem with the ResNet backbone and comparable performance with the Swin backbones. When we remove external data, \model substantially outperforms RichSem by $+6.0$ \apr in the fully supervised setting, implying that the success of RichSem is sensitive to the contributions of CLIP and ImageNet supervision. Our framework carries the benefit of being robust across settings with and without external data.

\begin{table}[t]
\centering
\resizebox{\linewidth}{!}{
    \begin{tabular}{lrr}
        \toprule
        Configuration &
        m\apbox &
        \apr\\
        \midrule \midrule
        Single-Stage Training +++Copy-Paste &
        28.1 &
        16.8\\
        \midrule
        Multi-Stage Training (w/ Random Resize) &
        26.8 &
        17.9\\
        Multi-Stage Training +Photometric Jittering &
        26.9 &
        18.2\\
        Multi-Stage Training ++Repeat Sampling &
        27.1 &
        19.3\\
        \cellcolor{Gray}Multi-Stage Training +++Copy-Paste (Ours) &
        \cellcolor{Gray}\textbf{29.0} &
        \cellcolor{Gray}\textbf{20.9}\\
        \bottomrule
    \end{tabular}
}
\captionof{table}{Ablation experiments quantifying the effectiveness of each component in our multi-stage training protocol. The single-stage model is trained end-to-end on the whole long-tailed dataset.}
\label{tab::ablation-supervised}
\end{table}

\subsection{Ablation Experiments}\label{sec::ablation}
\paragraph{Design of \model Baseline}
The design of \model is centered on an intuitive multi-stage training strategy combined with standard data augmentations, without bells and whistles, to overcome the class imbalances in both head and tail datasets. \Cref{tab::ablation-supervised} shows the contributions of RFS and Copy-Paste in establishing a more robust baseline than was previously possible in the existing literature, by using the simple Faster R-CNN detector with ResNet-50 FPN. We also emphasize the advantage of multi-stage learning over the na\"ive single-stage training procedure on the whole long-tailed dataset, which yields markedly worse results.

\begin{figure}[t]
  \centering
  \setlength\tabcolsep{1.0pt}
  \begin{tabular}{cc}
    \makecell{\includegraphics[height=0.34\linewidth]{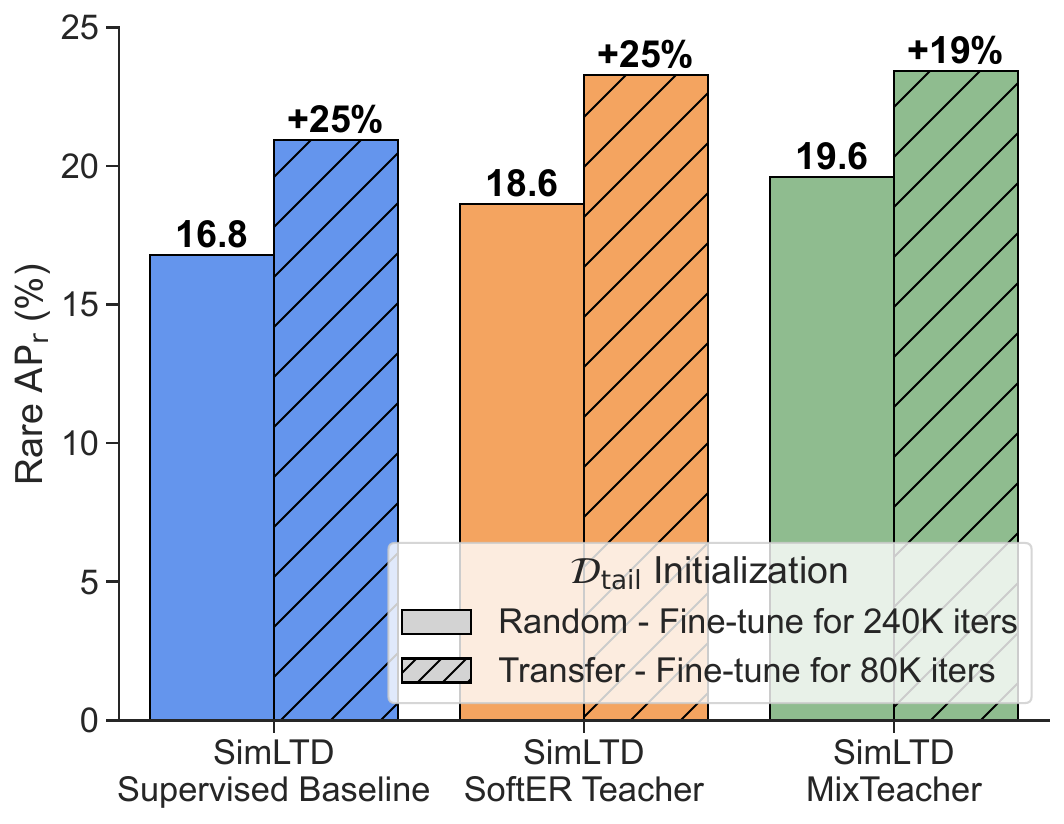}}&
    \makecell{\includegraphics[height=0.34\linewidth]{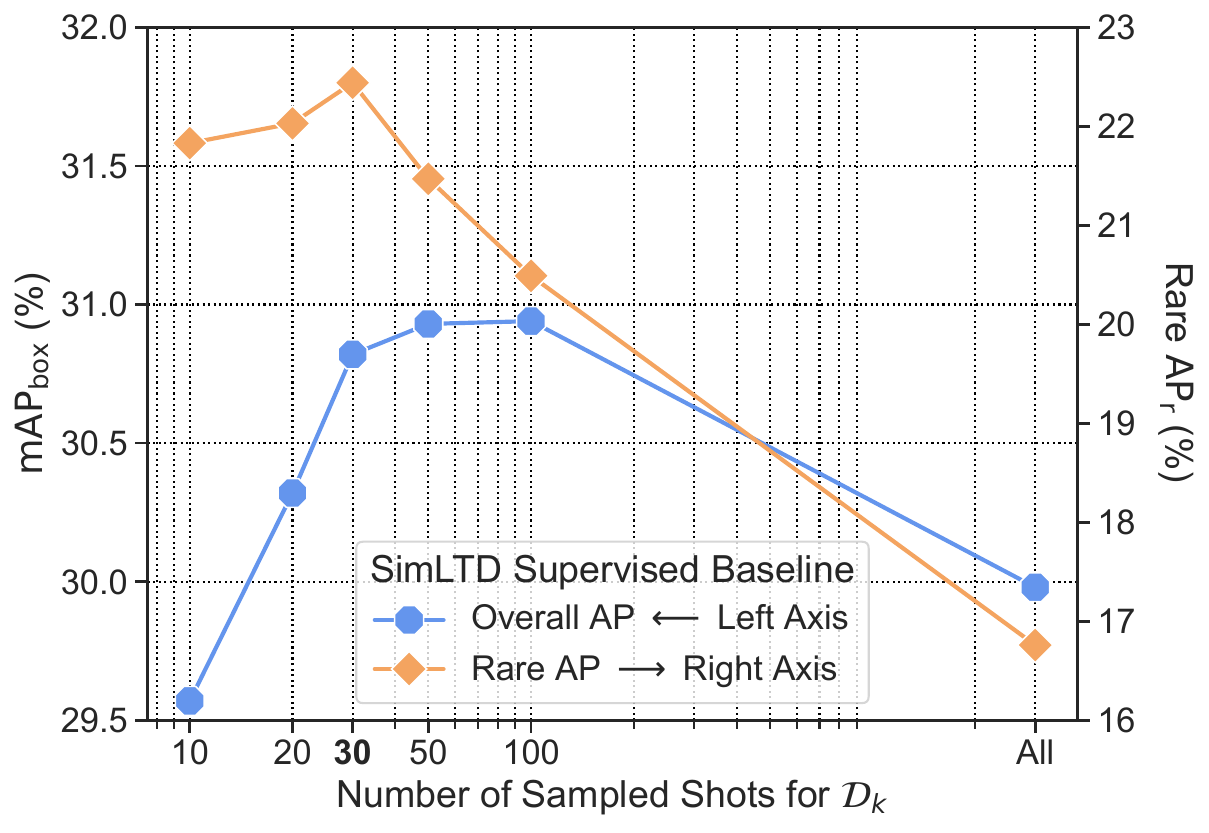}}\\
  \end{tabular}
  \vspace{-4pt}
  \caption{Ablation experiments assessing the impact on AP from transfer learning on tail classes \textbf{(left)} and from varying the number of sampled shots for fine-tuning with \dk \textbf{(right)}.}
  \label{fig::ablation-studies}
\end{figure}

\medskip
\noindent
\textbf{The Impact of Transfer Learning on} \dtail ~ 
As discussed in \S\ref{sec::step2}, we transfer the pre-trained head representations to optimize on tail classes in Step 2 of our framework. However, this step is optional and may be skipped, because we can initialize the tail detector with random weights before fine-tuning. \Cref{fig::ablation-studies} shows that transfer learning on tail classes is a worthwhile task. Across both supervised and semi-supervised settings, transfer learning gives a boost by up to $+4.7$ \apr and comes with the added bonus of shortening the fine-tuning time by $\nicefrac{2}{3}$ of the required iterations.

\medskip
\noindent
\textbf{How Many Shots to Sample for \dk?} ~
\Cref{fig::ablation-studies} illustrates the impact on m\apbox and \apr as a function of sampled shots for fine-tuning with \dk, ranging from 10 to ``All'' meaning the entire long-tailed training set. The aim of this experiment is to optimize for accuracy on rare classes while mitigating catastrophic forgetting on pre-trained head representations. We analyze the ``knee in the curve'' and find that 30-shots balance the trade-off between the two metrics. \Cref{fig::overview} visualizes that with 30-shots, we sample the whole tail distribution containing 20 or fewer instances per class and include a mixture of head categories for exemplar replay. Moving to the left of this ``sweet spot'' with $\{10,20\}$-shots, we observe marked reductions in m\apbox, indicating adverse forgetting on head classes from insufficient samples. Moving to the right of the sweet spot, we see a precipitous drop in \apr in response to the overwhelming amount of head samples.

\section{Conclusion}
We introduced \model, a versatile framework for effective supervised and semi-supervised long-tailed detection with unlabeled images. Standing out from existing work, \model delivers excellent performance and scalability to advance the challenging \lvis v1 object detection benchmark---without requiring auxiliary image-level supervision. We hope the practitioner finds utility in our multi-stage training approach and for our work to spur future research aimed at pushing the performance envelope of long-tailed detection.
\section*{Acknowledgments}
The author thanks Victor Palmer for the insightful discussion, Brian Keller, Andrew Melkonian, Mehran Ghandehari, Elan Ding, and anonymous reviewers for their thoughtful and constructive feedback on this paper.

{
  \small
  \bibliographystyle{ieeenat_fullname}
  \bibliography{main}
}

\end{document}